\documentclass[onecolumn,10pt]{IEEEtran}

\usepackage{amsmath,amsfonts,amsthm, amssymb, enumerate}
\usepackage{cite}
\usepackage{lipsum}
\usepackage[justification=centering]{caption}
\usepackage{amssymb,graphicx,subfigure}
\usepackage[bottom]{footmisc}
\usepackage[dvipsnames]{xcolor}
\usepackage{balance}

\usepackage{array,multirow}
\newlength{\Oldarrayrulewidth}

\usepackage{algorithm,algpseudocode}
\algnewcommand\Input{\item[{\textbf{Input:}}]}
\algnewcommand\Output{\item[{\textbf{Output:}}]}
\algnewcommand\Initialize{\item[{\textbf{Initialize:}}]}
\algnewcommand{\return}[1]{
  \State \textbf{return:}
  \Statex \hspace*{\algorithmicindent}\parbox[t]{.8\linewidth}{\raggedright #1}
}

\newcommand{\supp}{\operatorname{supp}}

\newcommand{\abs}{\operatorname{abs}}
\begin{document}
\title{Fourier Phase Retrieval with Extended Support Estimation via Deep Neural Network}

\author{Kyung-Su~Kim,~Sae-Young Chung\\
School of Electrical Engineering, Korea Advanced Institute of Science and Technology\\
E-mails: {kyungsukim, schung}@kaist.ac.kr}

\markboth{ }
{Shell \MakeLowercase{\textit{et al.}}: Bare Demo of IEEEtran.cls for Journals}

\maketitle
\begin{abstract} 
	We consider the problem of sparse phase retrieval from Fourier transform magnitudes to recover the $k$-sparse signal vector and its support $\mathcal{T}$. We exploit extended support estimate $\mathcal{E}$ with size larger than $k$ satisfying $\mathcal{E} \supseteq \mathcal{T}$ and obtained by a trained deep neural network (DNN). To make the DNN learnable, it provides $\mathcal{E}$ as the union of equivalent solutions of $\mathcal{T}$ by utilizing modulo Fourier invariances. Set $\mathcal{E}$ can be estimated with short running time via the DNN, and support $\mathcal{T}$ can be determined from the DNN output rather than from the full index set by applying hard thresholding to $\mathcal{E}$. Thus, the DNN-based extended support estimation improves the reconstruction performance of the signal with a low complexity burden dependent on $k$. Numerical results verify that the proposed scheme has a superior performance with lower complexity compared to local search-based greedy sparse phase retrieval and a state-of-the-art variant of the Fienup method.
\end{abstract}


\IEEEpeerreviewmaketitle

\section{Introduction} 
\label{intr}
Sparse phase retrieval from the magnitude of the Fourier transform (SPRF)~\cite{jaganathan2012recovery} has been widely studied in many fields including X-ray crystallography~\cite{millane1990phase}, optics~\cite{shechtman2015phase,katkovnik2017sparse}, blind channel estimation~\cite{baykal2004blind}, and computational biology~\cite{stefik1978inferring}. It recovers $k$-sparse\footnote{Signal vector ${x^{\circ}}$ is called $k$-sparse if it has $k$ nonzero elements.} signal vector ${x^{\circ}} = (x[1],...,x[n])^{\top}$ given measurements and the squared magnitude,  $y= (y[1],...,y[m])^{\top}$, of an $m$-point discrete Fourier transform of ${x^{\circ}}$:
\begin{align}\label{form1}
y[i] = c[i]+ w[i]  \,\, \textup{for}\,\,  i \in \{1,...,m\},
\end{align}
where $c[i] = \abs(\sum\limits_{j \in {\mathcal{T}}}  x[j] \exp{(- {2 \pi \sqrt{-1} (i-1)(j-1) }/{m})})^2$, $\abs(\cdot)$ denotes the  elementwise absolute value, ${\mathcal{T}}=\supp(x^{\circ})$ is the support of ${x^{\circ}}$ (i.e., set of nonzero elements in ${x^{\circ}}$) with size $k$, and $(w[1],...,w[m])^{\top} =: w \in \mathbb{R}^m$ is a noise vector. 

A commonly used algorithm to solve SPRF is the greedy sparse phase retrieval (GESPAR) proposed by Shechtman et al.~\cite{shechtman2014gespar}. 
GESPAR performs a local search for $\mathcal{T}$ and iteratively updates support estimate $\mathcal{S}$ by exchanging one element in $\mathcal{S}$  with one in $\mathcal{V} \setminus \mathcal{S}$, where $\mathcal{V}$ is an estimated index set such that $\mathcal{V} \supseteq \mathcal{T}$. Depending on the search technology, GESPAR exhibits better performance than related algorithms (e.g., sparse Fienup~\cite{mukherjee2012iterative}, SDP~\cite{jaganathan2012recovery}, and two-stage sparse phase retrieval~\cite{jaganathan2017sparse}) to reconstruct ${x^{\circ}}$. 
However, given that GESPAR updates only one index in the support estimate per iteration, its performance according to complexity (i.e., efficiency) can be severely degraded as the set difference between $\mathcal{S}$ and $\mathcal{T}$ widens, and its complexity scales with $k$~\cite{schniter2015compressive}. 
The complexity of GESPAR further increases as the signal dimension $n$ or the signal-to-noise ratio (SNR) increases, given that set $\mathcal{V}$ approaches the full index set, $\{1,...,n\}$, in any of these cases.

A learned deep neural network (DNN) can obtain desired solutions with high efficiency by simply performing a matrix multiplication at each layer without solving specific optimization problems. Therefore, the DNN has notably contributed to enhancing the performance of image reconstruction and denoising in SPRF \cite{goy2018low,metzler2018prdeep,rivenson2018phase}. However, this advantage has been limited to image processing, because DNNs consider image features during learning. Consequently, available research has neglected DNNs for performance improvement to recover general (synthetic) signals for SPRF. Nevertheless, verifying the high DNN performance for recovering any synthetic signal for SPRF would imply its superiority in all fields of SPRF besides image processing. On the other hand, DNNs have provided much lower complexity with similar performance to other algorithms to recover any synthetic signal in non-SPRF domains \cite{xin2016maximal,he2017bayesian,kim2019tree}. Hence, DNNs can improve the efficiency to recover all synthetic signals in the SPRF domain; we discussed this in detail in Section \ref{1discus}.
To verify this, we propose an algorithm called phase retrieval with extended support estimation using DNN (PRED). This is a one-shot retrieval for the support by exploiting the prior information via a DNN applied in SPRF. It improves the efficiency of GESPAR to recover all synthetic and sparse signals. In Section \ref{2discus}, we demonstrated PRED scalability through intuition and principles.

As long short-term memory (LSTM) has the same structure as Bayesian learning iterations~\cite{he2017bayesian}, the phase retrieval problem can be solved by using either the Bayesian learning framework~\cite{liu2018low} or implementing the framework as a subroutine~\cite{schniter2015compressive}. In fact, SPRF can be solved by executing the linear inversion for sparse estimation (e.g., sparse Bayesian learning) as a subroutine~\cite{schniter2015compressive}. 
Thus, LSTM-based DNNs implicitly enable to impose structural priors to estimate the support in SPRF. Therefore, we adopt a gated-feedback LSTM~\cite{he2017bayesian,chung2015gated} for the DNN in PRED, although other DNN architectures may be applied for SPRF.

PRED determines extended support estimate $\mathcal{E}$ to identify $\mathcal{T}$ (Section \ref{sec_f}). The extended support denotes an index set with size larger than sparsity $k$ and containing ${\mathcal{T}}$.   Specifically, we propose a DNN framework and its training rule to generate $\mathcal{E}$. For the DNN to be learnable, we define a union of equivalent solutions of $\mathcal{T}$ and train the DNN to estimate this union set instead of $\mathcal{T}$ (Section \ref{seca}). PRED iteratively obtains $\mathcal{E}$ from the trained DNN output and estimates $\mathcal{T}$ as a subset $\mathcal{S}$ of $\mathcal{E}$ through an algorithm called three-stage signal estimation (TSE); this process makes PRED scalable as we explained it in Section \ref{2discus}. TSE extends the damped Gauss--Newton (DGN) algorithm from GESPAR by taking more than $k$ indices as input (Section \ref{secb})~\cite{shechtman2014gespar}. In addition, PRED improves the efficiency of GESPAR to find $\mathcal{T}$. In fact, it simultaneously updates multiple indices in support estimate $\mathcal{S}$  by exploiting a probability measure for $\mathcal{E}$, which is provided by the trained DNN output, whereas GESPAR updates one index in $\mathcal{S}$ per iteration without utilizing the measure.
Numerical results confirm that PRED outperforms GESPAR and a state-of-the-art variant of the Fienup technique, called FISTA for phase retrieval (FISTAPH) \cite{pauwels2017fienup}, with lower complexity.

\section{Background}
\label{sec_b}

\subsection{DGN Algorithm}
Suppose that an estimate of ${\mathcal{T}}$ is given as ${\mathcal{S}}$ ($|{\mathcal{S}}|=k$, where $|\mathcal{S}|$  is the cardinality of $\mathcal{S}$). If ${\mathcal{S}}$ is correct (i.e., ${\mathcal{S}}={\mathcal{T}}$), SPRF can be formulated as the minimization in (\ref{b}), whose solution (i.e., $k$-sparse vector $ x \in \mathbb{R}^{n}$) is an estimate of $x^{\circ}$.
\begin{align}\label{b}
\begin{matrix}
\underset{ x \in \mathbb{R}^{n}}{\textup{minimize}} &g( x;\mathcal{S}):=\sum_{i\in \{1:m\}} ( y[i]- v_i( x;{\mathcal{S}}))^2, 
\end{matrix}
\end{align}
where $v_i({ x};{\mathcal{S}}):=({ x}^{\mathcal{S}})^{\top}A_i({\mathcal{S}}) \,{ x}^{\mathcal{S}}$, $A_i({\mathcal{S}}):=(F^{\{i\}}_{\mathcal{S}})^* F^{\{i\}}_{\mathcal{S}} \in \mathbb{R}^{k \times k}$  for $i \in \{1:m\}$, and $F \in \mathbb{C}^{m \times n}$ is the discrete Fourier transform such that $(c[1],...,c[m])^{\top} =:c$ can be expressed as $c=|Fx^{\circ}|^2$. For brevity, set $\{i,i+1,...,j\}$ is denoted as $\{i:j\}$. A submatrix of $A:=[a_1,...,a_n]$ $ \in \mathbb{R}^{m \times n}$ with columns indexed by $J \subseteq \{1:n\}$ is denoted by $A_J$. $A^{Q}$ and $x^{Q}$ denote the submatrix of $A$ with rows indexed by $Q \subseteq \{1:m\}$ and the subvector of $x$ with entries indexed by $Q$, respectively.

Using first-order linear approximation, $y[i] - v_i( x;{\mathcal{S}})$ in (\ref{b}) can be approximated as the $i$th element of vector $B( x_t;{\mathcal{S}})  x_t^{\mathcal{S}} - b( x_t;{\mathcal{S}})\in \mathbb{R}^{m}$, where $B( x_t;{\mathcal{S}}) \in \mathbb{R}^{m \times k }$ is the matrix whose $i$th row is $2(({ x_t}^{\mathcal{S}})^{\top} A_i({\mathcal{S}}))\in \mathbb{R}^{k}$ and $b( x_t;{\mathcal{S}})\in \mathbb{R}^{m }$ is the vector whose $i$th element is $y[i]+({ x_t}^{\mathcal{S}})^{\top} A_i({\mathcal{S}})\, { x_t}^{\mathcal{S}}$. Then, using ${\mathcal{S}}$, the DGN method applied in SPRF (Algorithm \ref{alg1}) estimates ${x^{\circ}}$ as a limit point of sequence $( x_1, x_2,...)$ obtained in steps 2 and 3, where $\delta_t:=(1/2)^{a(u, x_t,z_{t};\mathcal{S})} u$ is a step size determined by a backtracking procedure and $a(u, x_t,z_{t};\mathcal{S}) \in \mathbb{N}$ denotes the minimum nonnegative integer $a$ such that  $g( x_t -  \delta_t  d_t;\mathcal{S}) < g( x_t;\mathcal{S})-u(\frac{1}{2})^{a+1} \nabla g( x_t;\mathcal{S})^{\top}\, {d_t}^{\mathcal{S}}$. The limit of sequence $(x_{1}, x_{2},...)$ has been proven to be stationary~\cite{shechtman2014gespar}.
\begin{algorithm}[tb]
	\caption{DGN($F,y,\mathcal{S},\tau,h$)}
	\begin{algorithmic}[1]
		\Input{$\mathcal{S} \subseteq \{1:n\}$, $F \in \mathbb{R}^{m \times n}$, $h\in \mathbb{N}$, $\tau\in \mathbb{R}$}
		\Output{Signal estimate $\hat x  \in \mathbb{R}^{n}$}
		\Initialize{$ x_{1} \in \mathbb{R}^{n}$ is randomly generated.}
		\While{$\left \|  x_{t+1}- x_{t} \right \| > \tau$ or $t<h$ }
		\State $ (z_{t})^{\mathcal{S}}  \gets (B( x_t;{\mathcal{S}})^{\top}B( x_t;{\mathcal{S}}))^{-1}B( x_t;{\mathcal{S}})^{\top}b( x_t;{\mathcal{S}})$
		\State $ x_{t+1}  \gets  x_t - \delta_t d_t$, where $\delta_t:=(1/2)^{a(u, x_t,z_{t};\mathcal{S})} u$ and $d_t:=( x_t - z_{t})$
		\State $u \gets \min(2 \delta_t,1)$
		\State $t \gets t+1$
		\EndWhile
		\State \Return{$\hat x \in \mathbb{R}^{n}$ such that $(\hat x)^{\mathcal{S}} =  x_{t}$ and $(\hat x)^{\{1:n\}\setminus \mathcal{S}}=0$}
	\end{algorithmic}
	\label{alg1}
\end{algorithm}

\subsection{GESPAR}
GESPAR (Algorithm \ref{alg2}) first determines two index sets, $\mathcal{V}_1$ and $\mathcal{V}_2$, satisfying $\mathcal{V}_1 \subseteq \mathcal{T} \subseteq  \mathcal{V}_2$ through an autocorrelation-based process. Then, it generates estimate $\mathcal{S}$ of ${\mathcal{T}}$ ($|\mathcal{S}|=k$) such that $\mathcal{V}_1 \subseteq \mathcal{S} \subseteq  \mathcal{V}_2$ and utilizes the DGN method with $\mathcal{S}$ (Algorithm 1) to estimate signal ${x^{\circ}}$ and determine whether signal error $\epsilon$ is sufficiently small. GESPAR iteratively updates support estimate $\mathcal{S}$ in step 3 using 2-opt local search, and the iterative process (steps 2--11) proceeds until either the signal error is sufficiently small or GESPAR exceeds iteration limit $\kappa_{\textup{ITER}}$. More details on GESPAR can be found in~\cite{shechtman2014gespar}.

\begin{algorithm}[tb]
	\caption{GESPAR($F,y,k,\tau,h,\epsilon,\kappa_{\textup{ITER}}$)}
	\begin{algorithmic}[1]
		\Input{$(\mathcal{V}_1,\mathcal{V}_2) \subseteq \{1:n\}^2$, $F \in \mathbb{R}^{m \times n}$, $h\in \mathbb{N}$, $\tau\in \mathbb{R}$}
		\Output{Signal estimate $\hat x \in \mathbb{R}^n$, support estimate $\mathcal{S} \subseteq \{1:n\}$}
		\Initialize{Support estimate $\mathcal{S}_{1}$ of size $k$ is randomly generated such that $\mathcal{V}_1 \subseteq \mathcal{S}_{1} \subseteq \mathcal{V}_2$, where $\mathcal{V}_1$ and $\mathcal{V}_2$ are index sets obtained through the autocorrelation-based technique described in \cite{shechtman2014gespar}}.
		\State $x_{1} \gets \textup{DGN($F,y,\mathcal{S}_{1},\tau,h$)}$
		\For{$t=1 \textup{ to }  \kappa_{\textup{ITER}}$}
		\State $\mathcal{S}_{t+1} \gets$ Update support estimate $\mathcal{S}_t$ such that one index in $\mathcal{S}_t$ is replaced with one in $\mathcal{V}_2 \setminus (\mathcal{S}_t \cup \mathcal{V}_1)$ (i.e., apply 2-opt local search~\cite{shechtman2014gespar})
		\State $x_{t+1} \gets \textup{DGN($F,y,\mathcal{S}_{t+1},\tau,h$)}$
		\If {$g(x_t;{\mathcal{S}}_t) > g(x_{t+1};{\mathcal{S}}_{t+1}) $} 
		\State Update $\mathcal{S}_{t+1}$ by randomly sampling $k$ indices  satisfying $\mathcal{V}_1 \subseteq \mathcal{S}_{t+1} \subseteq \mathcal{V}_2$
		\State $x_{t+1} \gets \textup{DGN($F,y,\mathcal{S}_{t+1},\tau,h$)}$
		\EndIf
		\If {$g(x_{t+1};{\mathcal{S}}_{t+1}) \leq \epsilon$} go to step 12
		\EndIf
		\EndFor
		\State \Return{$(\hat x:=x_a, \mathcal{S}:=\mathcal{S}_{a})$ where $a:=\underset{a \in \{1:t\}}{\arg\min}\, g(x_t;{\mathcal{S}}_t)$}
	\end{algorithmic}
	\label{alg2}
\end{algorithm}

\section{PRED Structure}\label{sec_f}
To enhance the tradeoff between performance and complexity from GESPAR, the proposed PRED aims to determine extended support estimate ${\mathcal{E}} \subseteq \{1:n\}$ from a trained DNN output and recover ${x^{\circ}}$ via the proposed TSE using ${\mathcal{E}}$. 
Suppose that ${\mathcal{E}}$ is given and includes ${\mathcal{T}}$. Then, the SPRF problem can be formulated by (\ref{c}), which estimates ${x^{\circ}}$ as solution $ x \in \mathbb{R}^{n}$.
\begin{align}\label{c}
\begin{matrix}
\underset{ x}{\textup{minimize}} &g( x;\mathcal{E}) :=\sum_{i\in \{1:m\}} ( y[i] - v_i( x;{\mathcal{E}}))^2, \\ 
\textup{subject to} &|\supp( x)| \leq k,
\end{matrix}
\end{align}
where $v_i({ x};{\mathcal{E}}):=({ x}^{\mathcal{E}})^{\top}A_i({\mathcal{E}}) \,{ x}^{\mathcal{E}}$ for $i \in \{1:m\}$.  Note that the original SPRF problem is equal to (\ref{c}) with ${\mathcal{E}}$ replaced by $\{1:n\}$. Thus, by considering ${\mathcal{E}}$, the SPRF problem can be simplified such that the dimension of the target signal is reduced from $n$ to $|\mathcal{E}|$. Besides, it is easier to find ${\mathcal{E}}$ such that ${\mathcal{E}} \supseteq {\mathcal{T}}$ than to identify ${\mathcal{T}}$. Hence, PRED adopts this principle to generate ${\mathcal{E}}$ from a trained DNN output (Section \ref{seca}) and estimates ${x^{\circ}}$ by solving (\ref{c}) given ${\mathcal{E}}$ (Section \ref{secb}).

\subsection{DNN for Extended Support Estimation}\label{seca}
We propose a DNN structure and a training method to obtain extended support ${\mathcal{E}}$ of ${x^{\circ}}$. Given the trivial ambiguity in SPRF (modulo Fourier invariances \cite{jaganathan2017sparse}), there exists a $k$-sparse vector $x \in \mathbb{R}^{n}$ satisfying $\abs(Fx^{\circ})=\abs(Fx)$, whose support $\mathcal{W}$ is defined by $\mathcal{T}- \min\limits_{v\in \mathcal{T}}v+l$ or $\max\limits_{v\in \mathcal{T}}v -\mathcal{T}+l$  for any integer $l$ such that $1\leq l \leq n -\max\limits_{v\in \mathcal{T}}v+\min\limits_{v\in \mathcal{T}}v$. 
Therefore, even in the noiseless case, support $\mathcal{W}$ of $x$ cannot be uniquely identified as $\mathcal{T}$ given the measurement vector $y$, and consequently it is hard to optimize the DNN for estimating $\mathcal{T}$. To solve this problem, we introduce union $\mathcal{I}(\mathcal{T})$  of equivalent solutions of the true support (UES), defined as $\mathcal{I}(\mathcal{T}):= \alpha(\mathcal{T}) \cup \beta(\mathcal{T})$, 
where $\alpha(\mathcal{T}):=\mathcal{T}- \min\limits_{v\in \mathcal{T}}v+1$, $\beta(\mathcal{T}):=\max\limits_{v\in \mathcal{T}}v-\mathcal{T}+1$. 
Unlike true support $\mathcal{T}$, UES $\mathcal{I}(\mathcal{T})$ is uniquely determined by $y$. Thus, the DNN is learnable without the ambiguity by considering its output as $\mathcal{I}(\mathcal{T})$ instead of $\mathcal{T}$.
As the index $1$ is always included in $\mathcal{I}(\mathcal{T})$, the DNN is trained to retrieve $\mathcal{I}_{-1}(\mathcal{T}):=\mathcal{I}(\mathcal{T}) \setminus \{1\}$.

\subsubsection{DNN Structure and Training Objective}
\label{nns}

For sparse vector $z \in \mathbb{R}^{n}$, whose support is denoted by $\mathcal{T}_z$ and measurement vector $h=\abs(F z)^2 \in \mathbb{R}^{m}$ given discrete Fourier transform matrix $F$, the DNN defined by $f_{\theta}(\cdot):\mathbb{R}^{m} \rightarrow \mathbb{T}^{n-1}$ takes vector $h$ as its input and is trained to return vector $v =(v[1],...,v[n-1])^{\top}= f_{\theta}(h) \in \mathbb{T}^{n-1}$ such that $v[i-1] = 1/|\mathcal{I}_{-1}(\mathcal{T}_z)|$ for $i \in \mathcal{I}_{-1}(\mathcal{T}_z)$ and $v[i-1] = 0$ for $i \notin \mathcal{I}_{-1}(\mathcal{T}_z)$, where $\mathbb{T}^q$ represents the $(q-1)$-dimensional probability simplex. Each element $v[i]$ of vector $v$ is a probability, with index $i+1$ belonging to $\mathcal{I}_{-1}(\mathcal{T}_z)$. Thus, for any integer $e$ such that $|\mathcal{I}_{-1}(\mathcal{T}_z)|\leq e$, an extended support $\mathcal{E}$ of $z$ including $\mathcal{I}_{-1}(\mathcal{T}_z)$ can be obtained by selecting the $e$ largest elements from the ideally trained DNN output. For instance, if $n$ is 6 and support $\mathcal{T}_z$ is $\{2,3,6\}$ so that $\mathcal{I}_{-1}(\mathcal{T}_z)=\{2,4,5\}$, $f_{\theta}(h)$ is
trained to return output vector $f_{\theta}(h)$ as $(1/3,0,1/3,1/3,0)^{\top}$, and $\mathcal{I}_{-1}(\mathcal{T}_z)$ is obtained by selecting its $3$ largest elements.

\subsubsection{DNN Training}

We randomly sampled a $k$-sparse signal vector $\bar x$, whose sparsity $k$ ranges from $k_1$ to $k_2$. {We set $(k_1,k_2)$ to $(2,20)$ in this study (Section \ref{sec_sim}). From signal vector $\bar x$, a measurement vector $\bar y$ satisfying (\ref{form1}) can be obtained by adding random noise vector $w$ according to the given SNR. For the distribution of pairs $(\bar x,\bar y)$ produced this way, the training goal can be formulated as the minimization of (\ref{costf}).
	\begin{align}\label{costf}
	L(\theta):= \mathbb{E}_{(\bar y,\bar x)} [ \textup{ce(}f_{\theta}(\bar y),\textup{u}_n(\mathcal{I}_{-1}(\mathcal{T}_{\bar x}) -1) ],
	\end{align} 
where $\textup{ce}(v_1,v_2):= -\frac{1}{g} \sum_{i=1}^{g} v_2[i] \log \, v_1[i] $ is the cross-entropy between vectors $v_1$ and $v_2$ of dimension $g$, $\textup{u}_n(\mathcal{I})$ is an $(n-1)$-dimensional vector $v \in \mathbb{R}^{n-1}$, whose nonzero elements are $1/|\mathcal{I}|$ and its support is given by set $\mathcal{I}$, $\mathcal{T}_{\bar x}$ is the support of ${\bar x}$, and $\theta$ is the training parameter to be updated for minimizing $L(\theta)$.  
The detailed description is shown in Appendix.

\subsection{Three-Stage Signal Estimation}
\label{secb}
\begin{algorithm}[tb]
	\caption{TSE(${F},y,k,\mathcal{E},\tau,h$)}
	\begin{algorithmic}[1]
		\Input{${F} \in \mathbb{C}^{m \times n}$, $y \in \mathbb{R}^n$, $\mathcal{E} \subseteq \{1:n\}$, $(k,h) \in \mathbb{N}^3$, $\tau \in \mathbb{R}$}
		\Output{Signal estimate $\hat x \in \mathbb{R}^n$, support estimate $\mathcal{S} \subseteq \{1:n\}$}
		\State $\bar x \gets \textup{DGN($F,y,\mathcal{E},\tau,h$)}$
		\State ${\mathcal{S}} \gets \{ \textup{indices of the $k$ largest values of $\abs(\bar x[z])$ for $z \in \{1:n\}$}\}$
		\State $\hat x \gets \textup{DGN($F,y,\mathcal{S},\tau,h$)}$
		\State  \Return{$(\hat x,\mathcal{S})$}
	\end{algorithmic}
	\label{alg4}
\end{algorithm}
From the trained DNN output, we can obtain extended support estimate $\mathcal{E}$ such that ${\mathcal{E}} \supseteq {\mathcal{I}(\mathcal{T})}$. Then, it remains to estimate the true support as $\alpha(\mathcal{T})$ or $\beta(\mathcal{T})$ by selecting $k$ indices from $\mathcal{E}$. This is resolved by the minimization in (\ref{c}) given ${\mathcal{E}} \supseteq {\mathcal{I}(\mathcal{T})}$. We introduce the TSE in Algorithm \ref{alg4}, which calls the DGN method (Algorithm \ref{alg1}) twice and approximately solves the minimization in (\ref{c}) through the following three stages: (a) temporary signal estimation from the given $\mathcal{E}$: $\textup{$\bar x \gets \underset{x \in \mathbb{R}^{n}}{\arg\min} \, g(x;\mathcal{E}) $}$; (b) support estimation from $\bar x$:  $\textup{$\mathcal{S} \gets \underset{\mathcal{S} \subseteq \mathcal{E} \textup{ s.t. }|\mathcal{S}|=k}{\arg\min} \, g(\bar x;\mathcal{S}) $}$; and (c) signal estimation from $\mathcal{S}$: $\textup{$\hat x\gets \underset{\check  x \in \mathbb{R}^{n}}{\arg\min} \, g(\check x;\mathcal{S}) $}$, where $\hat x$ and $\mathcal{S}$ are the signal and support estimates, respectively.

The first stage of TSE (step 1) minimizes the cost in (a), with a temporary signal estimate $\tilde x$ supported on ${\mathcal{E}}$ obtained by applying the DGN method (Algorithm \ref{alg1}) with ${\mathcal{E}}$ such that $\tilde x=$ DGN($F,y,\mathcal{E},\tau,h$). The second stage (step 2) retrieves support estimate ${\mathcal{S}}$ as a subset of $\mathcal{E}$ by approximating the minimization in (b) through hard thresholding (i.e., selecting $k$ indices corresponding to the $k$ largest absolute values of $(\tilde x)^{\mathcal{E}}$ supported on ${\mathcal{E}}$). Finally, TSE determines ${x^{\circ}}$ through the DGN method with ${\mathcal{S}}$ to solve (c) at the third stage (step 3).

\begin{algorithm}[t]
	\caption{PRED(${F},y,k,\tau,h,\kappa_{\textup{ITER}}$)}
	\begin{algorithmic}[1]
		\Input{${F} \in \mathbb{C}^{m \times n}$, $y \in \mathbb{R}^n$, $(k,h,\kappa_{\textup{ITER}}) \in \mathbb{N}^3$, $\tau \in \mathbb{R}$, trained neural network $f_{\theta}(\cdot): \mathbb{R}^{m} \rightarrow \mathbb{T}^{n-1}$}
		\Output{Signal estimate $\hat x \in \mathbb{R}^n$, support estimate $\mathcal{S} \subseteq \{1:n\}$}
		\Initialize{Integer $q$ is uniformly sampled in the interval from $2k$ to $3k$. Index set ${\mathcal{E}_1}$ of size $c$ is obtained as $\{1\} \cup \{$integers $i+1$ corresponding to the $(q-1)$ largest values of $d[i]$ for $i \in \{1:n-1\}\}$, where $d:=(d[1],d[2],...,d[n-1])^{\top}$ denotes the vector with $d[i]$ being the $i$th value of $f_{\theta}(y)$ for $i\in \{1:n-1\}$.} Iteration counter $t$ is set to $1$.
		
		\While{$t \leq  \kappa_{\textup{ITER}}$}
		\State $(v,{\mathcal{S}}_t,x_t) \gets \textup{TSE}({F},y,k,\mathcal{E}_t,\tau,h)$
		\If {$\sum_{i\in \{1:m\}} \abs( y[i] - v_i(x_t;\mathcal{S}_t) ) \leq \epsilon$}  go to step 10
		\EndIf
		\State Obtain discrete probability vector $p:=v/\left \| v\right \| \in \mathbb{R}^{n-1}$, where $v:=(v[1],...,v[n-1])^{\top}$ is a vector such that $v[i]=0$ for $i \in {\mathcal{S}}_t$ and $v[i]=d[i]$ for $i \in \{1:n-1\} \setminus {\mathcal{S}}_t$
		\State Sample integer $q$ from uniform distribution between $2k$ and $3k$
		\State ${\mathcal{E}_{t+1}} \gets$   $\{1\}\cup {\mathcal{S}}_t \cup {(\mathcal{G}+1)}$ where $\mathcal{G}$ is a set of $q-k-1$ indices randomly sampled without replacement according to the probability $p$
		\State  $t \gets t+1$
		\EndWhile
		\State \Return{$(\hat x := x_h, \mathcal{S} := {\mathcal{S}}_h)$, where $h:=\underset{q \in \{1:t\}}{\arg\min} \,\, g(x_q;{\mathcal{S}}_q) $}
	\end{algorithmic}
	\label{alg5}
\end{algorithm}
\section{PRED Algorithm}\label{sec_ai} 
The proposed PRED is detailed in Algorithm \ref{alg5} and estimates $({x^{\circ}},{\mathcal{T}})$ as $({\hat x},{\mathcal{S}})$ through trained DNN $f_{\theta}(\cdot)$ and by applying TSE (Algorithm \ref{alg4}). The trivial ambiguity of SPRF~\cite{jaganathan2017sparse} implies that index $1$ can be considered an element in the true support. By selecting the $(q-1)$ largest values of the trained DNN output, PRED initializes extended support estimate $\mathcal{E}$, where $q$ is the size of $\mathcal{E}$ randomly sampled from $2k$ to $3k$. Note that $|\mathcal{E}|$ is larger than $|\mathcal{I}(\mathcal{T})|$ from inequality $|\mathcal{E}| \geq 2k$, and the $i$th element of the trained DNN output $f_{\theta}(y)\in \mathbb{R}^{n-1}$ indicates a probability of index $i+1$ belonging to UES $\mathcal{I}(\mathcal{T})$. Thus, $\mathcal{E}$ is the set expected to include one of the equivalent solutions of $\mathcal{T}$ in $\mathcal{I}(\mathcal{T})$ (i.e., either $\alpha(\mathcal{T})$ or $\beta(\mathcal{T})$). Under premise $\mathcal{E} \supseteq \alpha(\mathcal{T})$ or $\mathcal{E} \supseteq \beta(\mathcal{T})$, PRED solves the minimization in (\ref{c}) by executing TSE to estimate $(x^{\circ},\mathcal{T})$ as ($x_1,\mathcal{S}_1$) in step 2. Then, PRED terminates depending on whether current signal error $g(x_1;\mathcal{S}_1)$ obtained from the estimate is below input threshold $\epsilon$ in steps 3 and 4.  If the signal error is higher than $\epsilon$, PRED executes steps 5--7 to obtain a new extended support estimate $\mathcal{E}$ via an update process, which is executed in step 7 by replacing the complementary set of $\mathcal{S}$ in $\mathcal{E}$ with multiple indices randomly selected according to discrete probability vector $p$ generated from DNN output $f_{\theta}(y)$. Vector $p$ represents the probability of each index in $\{2:n\} \setminus \mathcal{S}$ belonging to $\mathcal{I}(\mathcal{T}) \setminus \mathcal{S}$. By using the updated extended support $\mathcal{E}$, TSE estimates the signal vector and its support as ($x_2,\mathcal{S}_2$) in step 2 at the next iteration. This process is iterated until either the signal error is sufficiently small or the number of iterations exceeds limit $\kappa_{\textup{ITER}}$.

\section{Numerical Experiments and Results}
\label{sec_sim}
We compared the performance of PRED against similar algorithms, namely, GESPAR, FISTAPH, and phase-retrieval generalized approximate message passing (PRGAMP) \cite{schniter2015compressive}. For a fair comparison, we applied the same stopping criterion shown in steps 3 and 4 of Algorithm \ref{alg5} to GESPAR and FISTAPH. We executed FISTAPH $20i$ times for $n=256 (4-i)$ and $i\in \{1:3\}$ to get multiple candidate solutions and select the one with minimum error among them. The soft thresholding parameter of FISTAPH was set to $0.02$, and we selected the $k$ largest elements of its signal estimate to recover support $\mathcal{T}$. We used uniform and Gaussian models to generate signals. In the uniform model, each nonzero element of ${x^{\circ}}$ was sampled from $-1$ to $1$ excluding the interval from $-0.2$ to $0.2$. In the Gaussian model, each nonzero element of ${x^{\circ}}$ was sampled from a standard Gaussian distribution.  We assumed that each entry of $w$ follows chi-squared distribution $\chi^2(2)$ with $2$ degrees of freedom {For a complex random variable $z:=a+ b\sqrt{-1} \in \mathbb{C}$, whose real and imaginary parts are i.i.d following a Gaussian distribution, $|  \sum\limits_{j \in {\mathcal{T}}}  x[j] \exp{(- {2 \pi \sqrt{-1} (i-1)(j-1) }/{m})} +z|^2 \approx c[i] + a^2 +b^2$. Hence, the $i$th element of noise  vector $w$ can be set to $a^2 +b^2$,} such that $w[i] \sim \sigma \cdot \chi^2(2)$ for $i \in \{1:m\}$, where $\sigma$ is determined by the SNR. Given support estimate $\mathcal{S}$, we use modulo Fourier invariances and define recovery success rate $\mathbb{E}[\max$$(1(\alpha(\mathcal{T})=\alpha(\mathcal{S})),1(\alpha(\mathcal{T})=\beta(\mathcal{S})))]$ and soft recovery success rate $\mathbb{E}[\max$$(|\alpha(\mathcal{T})\cap\alpha(\mathcal{S})|,|\alpha(\mathcal{T})\cap \beta(\mathcal{S})|)/k]$ for the support, where $1(\cdot)$ is the indicator function.  We set  input parameters $(h,\tau,\epsilon) $ in PRED and GESPAR to $(100,10^{-4},\left \| y \right \| \cdot 10^{-v_{\textup{SNR}_{\textup{dB}}}/20})$ and iteration limit $\kappa_{\textup{ITER}}$ in PRED and GESPAR to $100$ and $500$, respectively.

For the gated-feedback LSTM  $f_\theta(\cdot)$ used in PRED~\cite{chung2015gated}, we set the hidden unit size, number of unfolding steps, and layer size to $2000$, $20$, and $2$, respectively. Further details on this network are available in~\cite{he2017bayesian}.
The detailed description and settings for learning DNN $f_\theta(\cdot)$ are given in Appendix.

\begin{figure*} 
	\centering
	\footnotesize
	\subfigure[\label{fig1a}(SNR,n) = (30,256)]{\includegraphics[width=2.9cm, height=2.40cm]{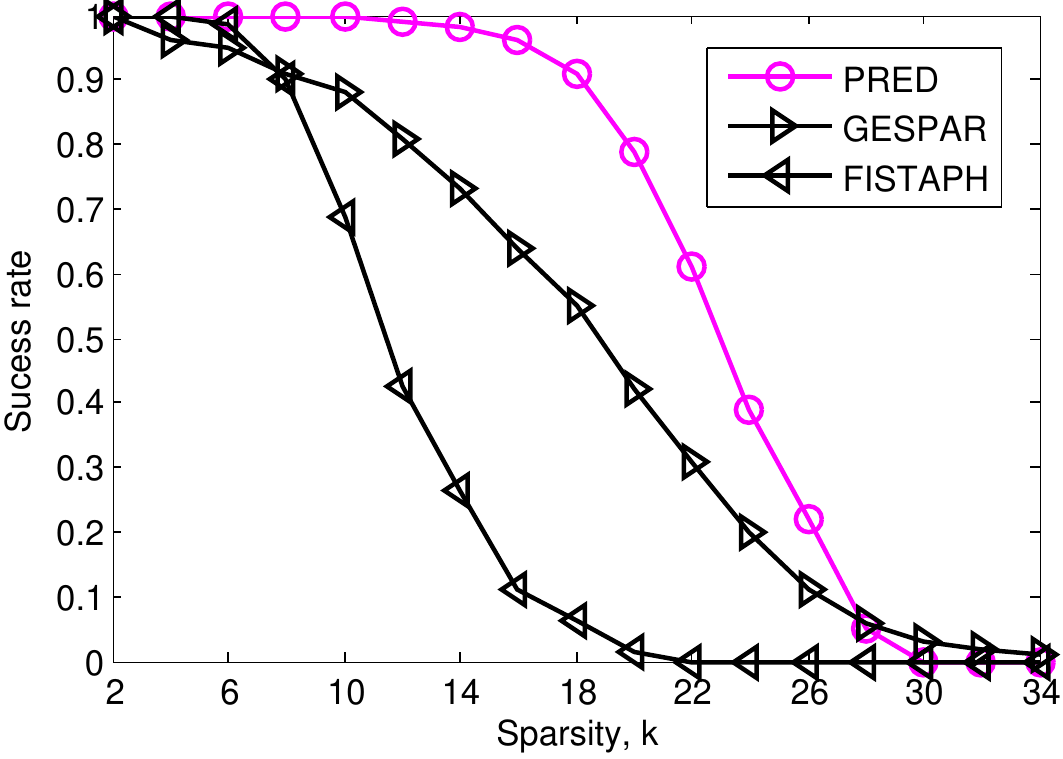}}
	\subfigure[\label{fig1b}(SNR,n) = (30,512)]{\includegraphics[width=2.9cm, height=2.40cm]{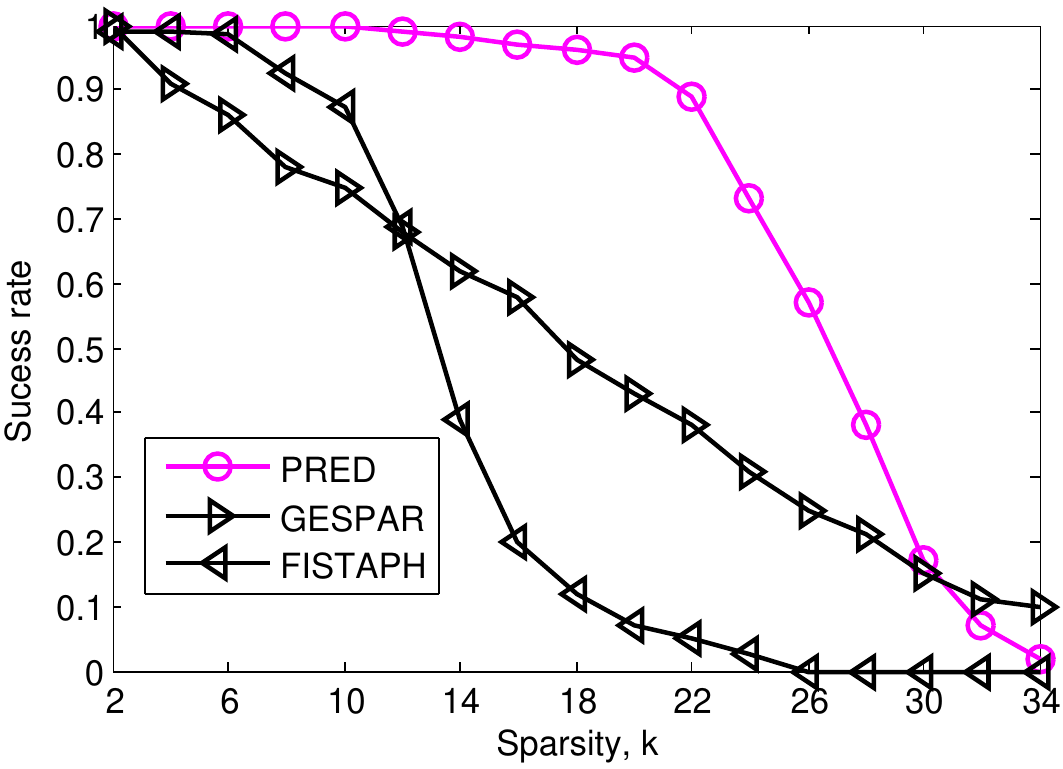}}
	\subfigure[\label{fig1c}(SNR,n) = (30,768)]{\includegraphics[width=2.9cm, height=2.40cm]{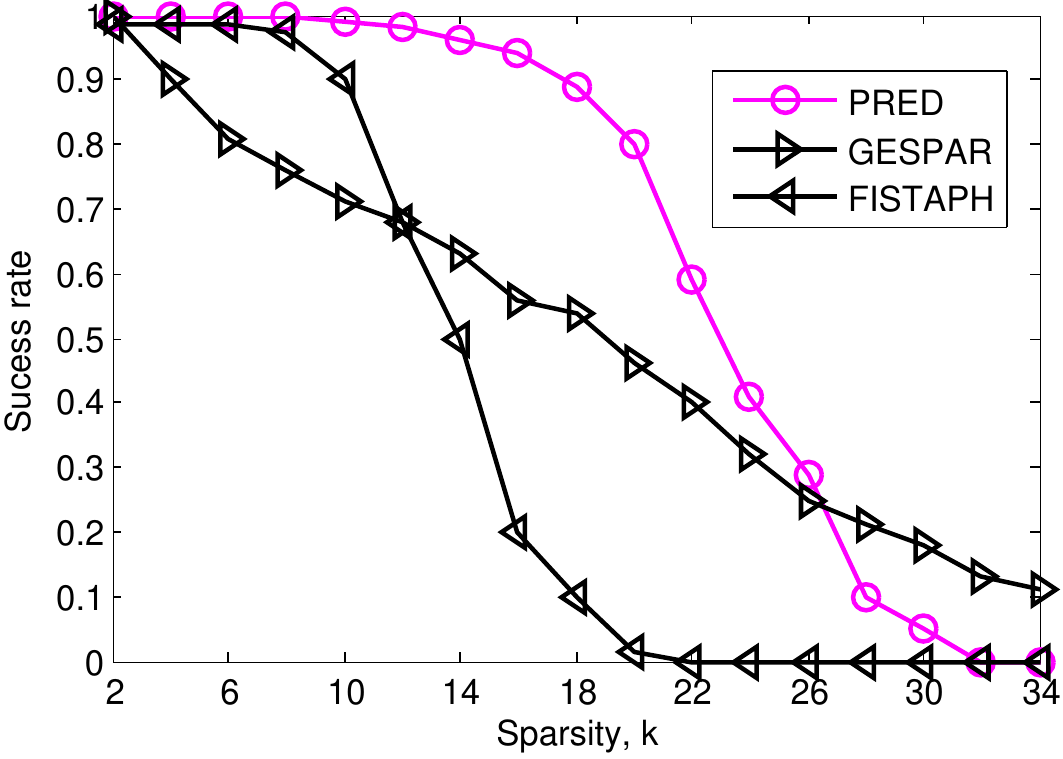}}
	\subfigure[\label{fig1d}(SNR,n) = (15,768)]{\includegraphics[width=2.9cm, height=2.40cm]{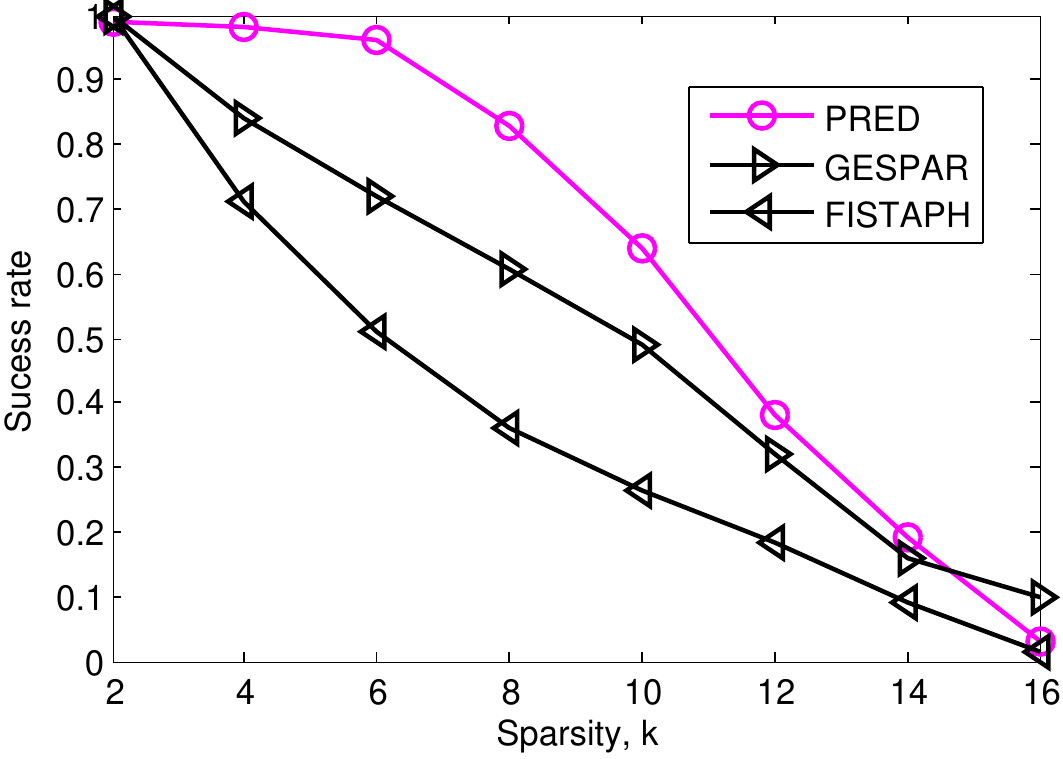}}
	\subfigure[\label{fig1e}(SNR,n) = (30,256)]{\includegraphics[width=2.9cm, height=2.40cm]{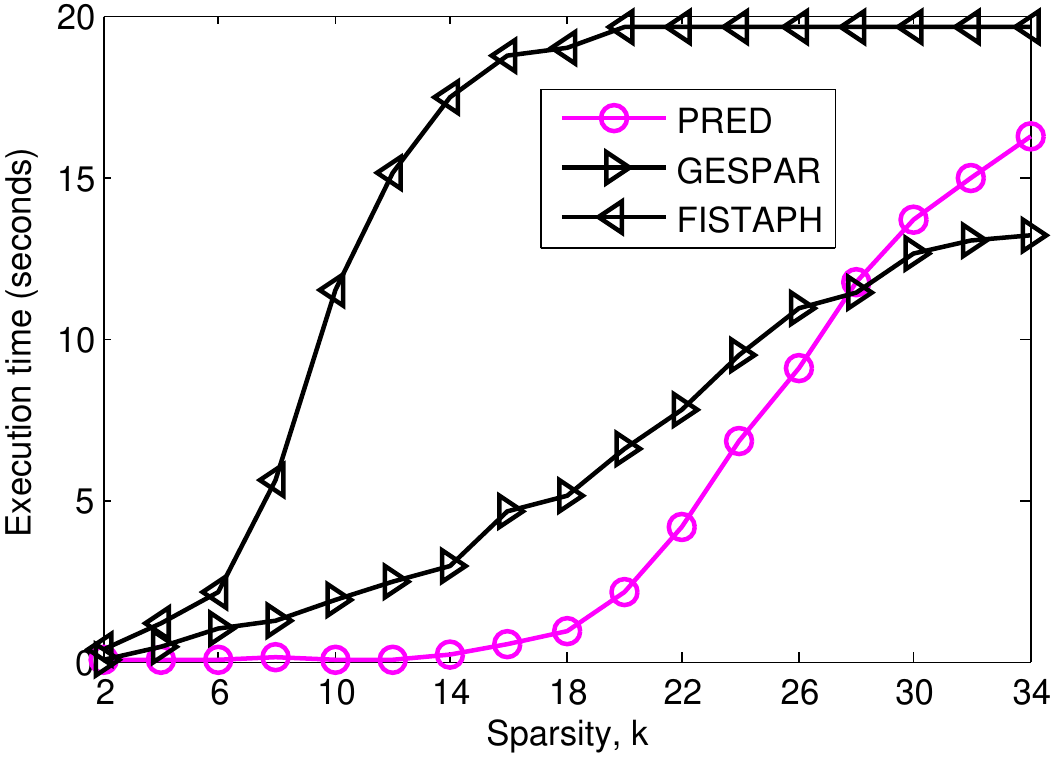}}
	\subfigure[\label{fig1f}(SNR,n) = (30,768)]{\includegraphics[width=2.9cm, height=2.40cm]{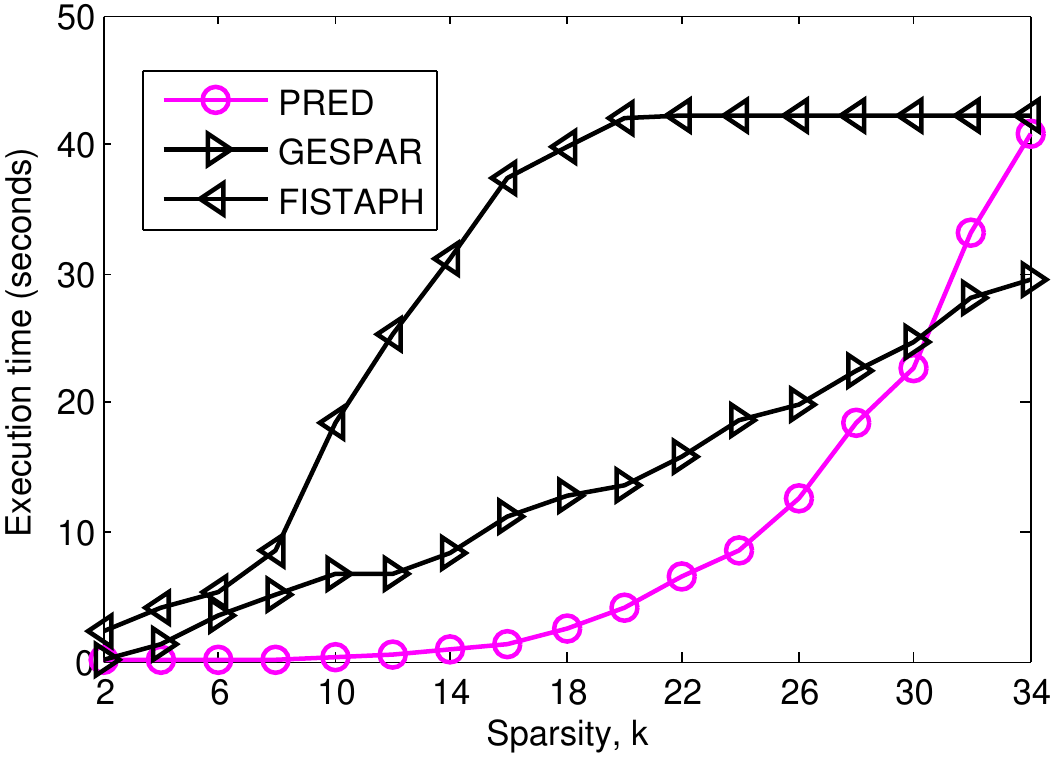}}
	\subfigure[\label{fig1g}(SNR,n) = (15,768)]{\includegraphics[width=2.9cm, height=2.40cm]{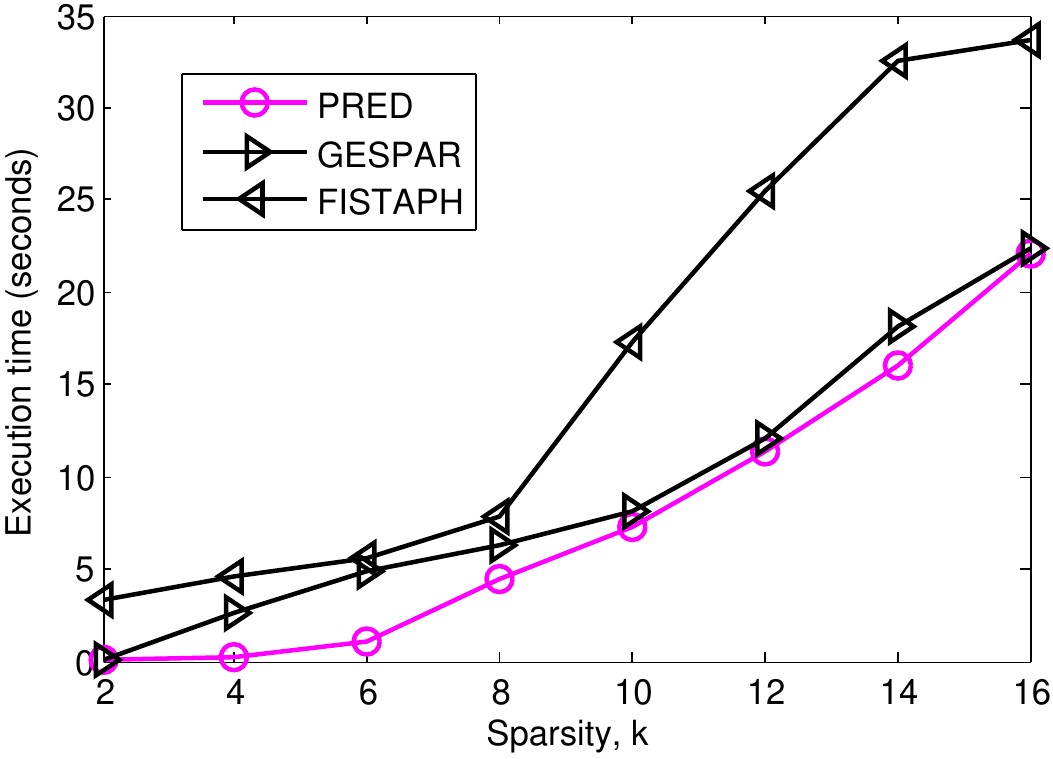}}
	\subfigure[\label{fig1h}(SNR,n) = (30,768)]{\includegraphics[width=2.9cm, height=2.40cm]{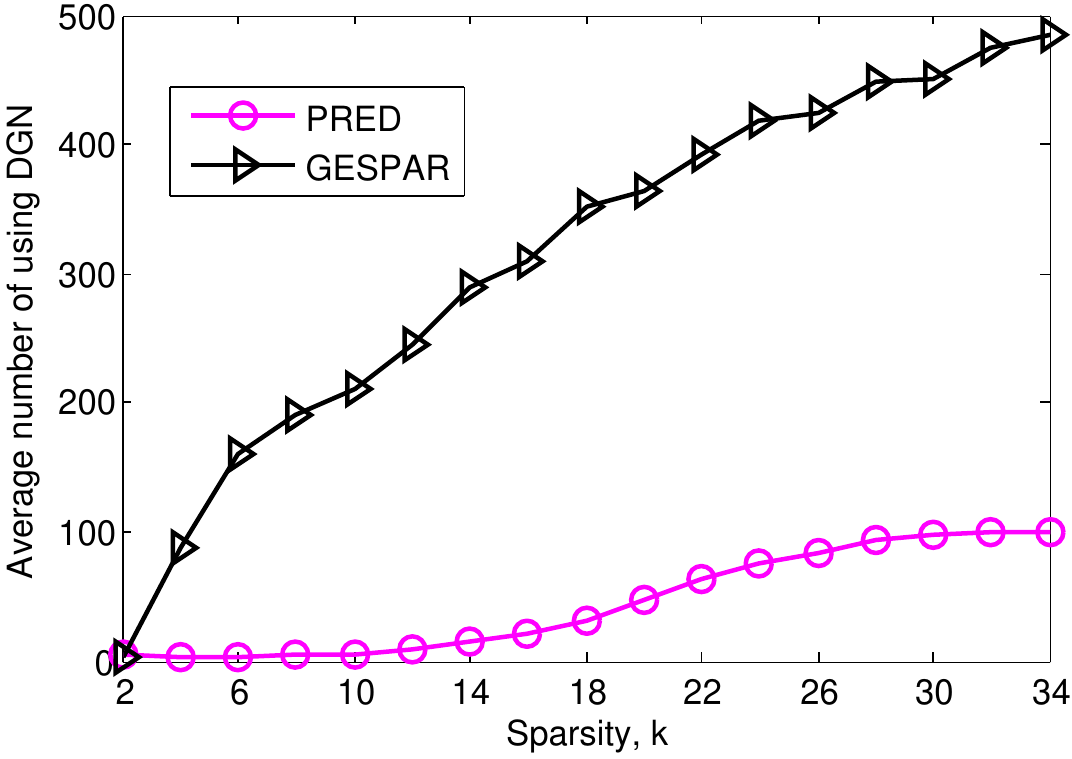}}
	\subfigure[\label{fig1i}(SNR,n) = (15,768)]{\includegraphics[width=2.9cm, height=2.40cm]{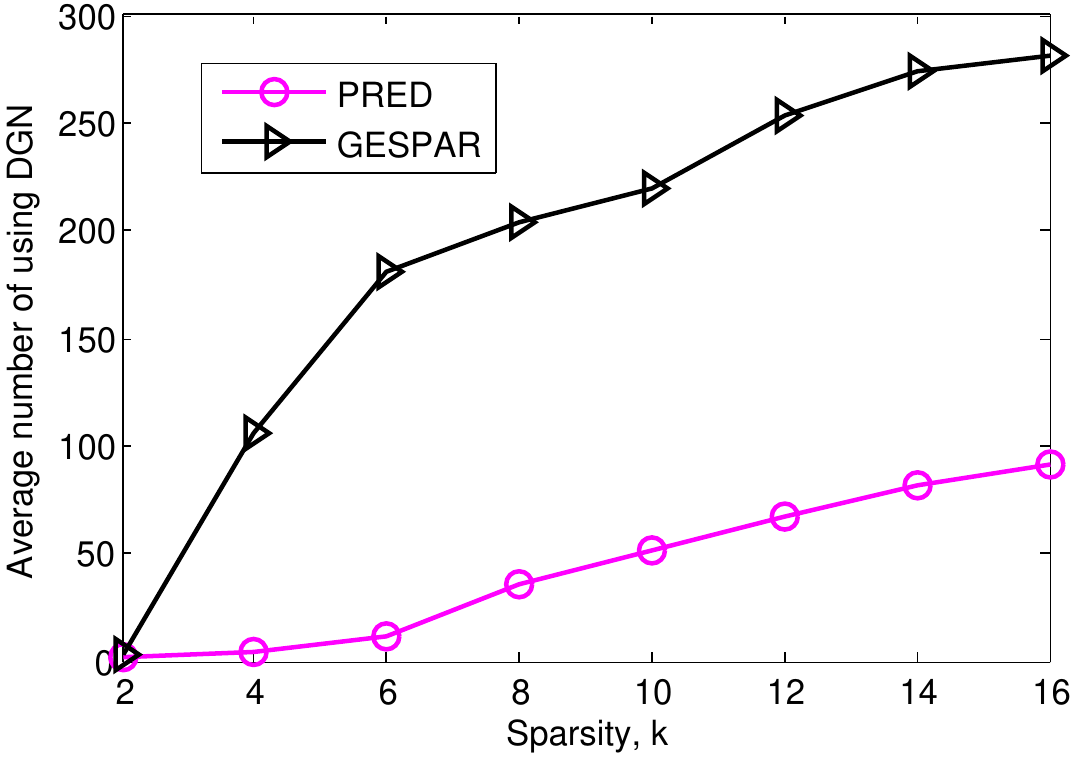}}
	\subfigure[\label{fig1j}(SNR,n) = (30,512)]{\includegraphics[width=2.9cm, height=2.40cm]{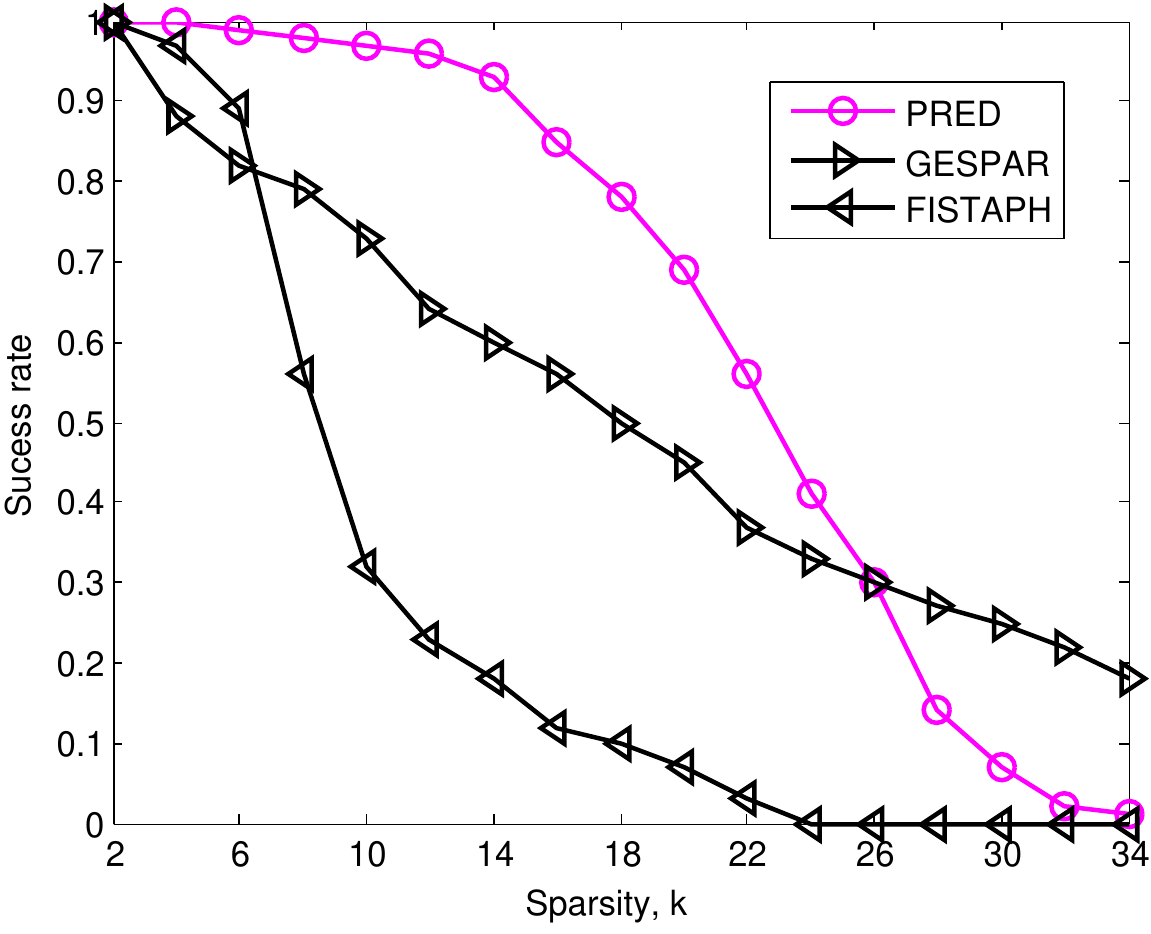}}
	\subfigure[\label{fig1k}(SNR,n) = (30,512)]{\includegraphics[width=2.9cm, height=2.40cm]{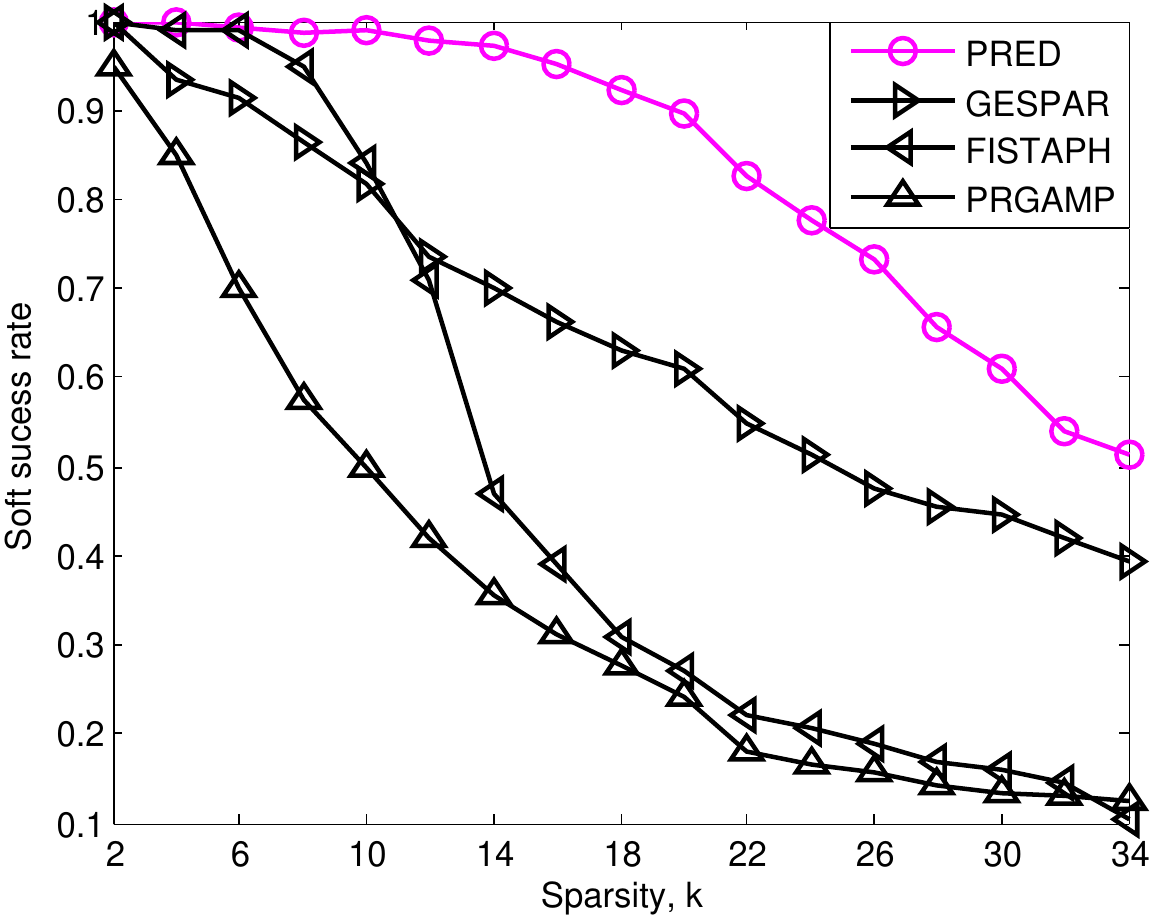}}
	\subfigure[\label{fig1l}(SNR,n) = (30,512)]{\includegraphics[width=2.9cm, height=2.42cm]{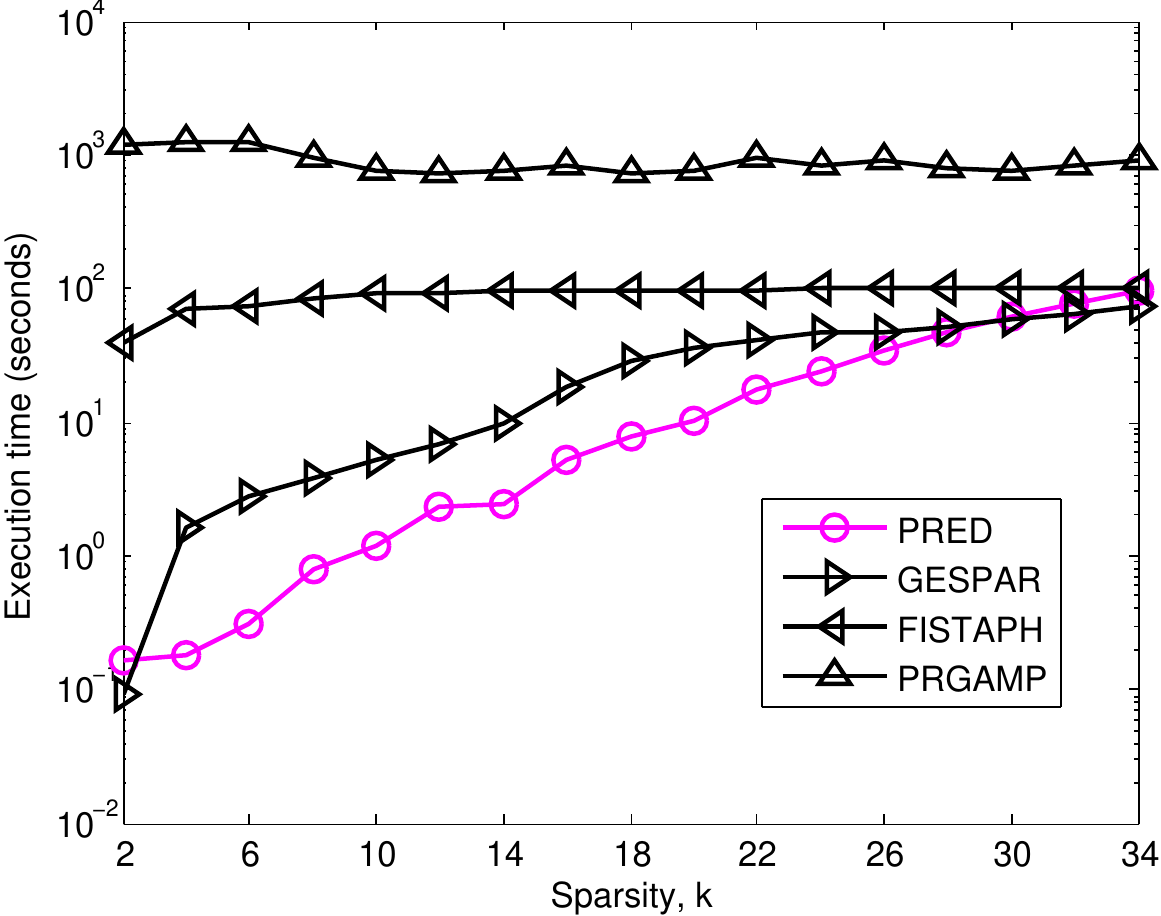}}
	\caption{Performance result according to SNR and pair ($m,n$) such that $m=n+1$. For the uniform model: (a)--(d) support recovery rate, (e)--(g) running time, and (h)--(i) average number $\eta$ of DGN  executions.  For the Gaussian model: (j) support recovery rate, (k) soft support recovery rate, and (l) running time.}
	\label{fig1}
\end{figure*}

We evaluated each algorithm with different SNRs and dimension values $n$ of ${x^{\circ}}$. Figs.~\ref{fig1a}--(d) and Figs.~\ref{fig1e}--(g) show the rate of successful support recovery and execution time per algorithm, respectively. In most of the sparsity region, PRED outperforms the other algorithms, provides lower complexity, and is more robust to noise. We can expect that PRED scales well with $n$, as Figs. 1(a)--(d) show that PRED uniformly recovers about twice the sparsity compared to GESPAR and FISTAPH for different $n$ and SNR values.\footnote{In Fig. 1(b), the maximum $k$ satisfying support recovery rates higher than $95$\% is $16$, $2$, and $8$ for PRED,  GESPAR, and FISTAPH, respectively.} Figs. 1(e)--(f) show that the running time of PRED is less than half of that of the other methods at sparsity $k$ below $20$.

Given that PRED and GESPAR consist of an iteration of DGN (Algorithm \ref{alg1}), their complexity is expressed as $\eta \cdot \phi$, where $\phi$ is the average complexity of DGN and $\eta$ is the average number of DGN executions in each algorithm. Note that the complexity of the pseudoinverse in step 2 of DGN with input $\mathcal{S}$ of size $b$ is generally $O(b^2 \cdot m)$, where $b$ is set to $k$ and $\alpha k$ in GESPAR and PRED, respectively, for a constant $\alpha$ smaller than $3$.\footnote{Given that size $b$ of the index set for the DGN input in steps 1 and 3 in TSE (Algorithm \ref{alg4}) is smaller than $3k$ and equal to $k$, respectively, $\alpha$ and its mean are smaller than $3$ and $3/2$, respectively.} This implies that average complexity $\phi$ of the DGN used in PRED and GESPAR has the same order for $k$, and hence their complexity is mainly dependent on $\eta$. Figs. \ref{fig1h}--(i) show that $\eta$ in PRED is less than or similar to one-third of $\eta$ in GESPAR. Thus, the complexity of PRED and GESPAR has the same order for $k$ and supports the results in Figs.~\ref{fig1e}--(g), showing that the execution time of PRED is shorter than that of GESPAR in most of the sparsity region. 

Figs.~\ref{fig1j}--(l) show the performance results for the zero-mean and unit-variance Gaussian model. PRGAMP\footnote{The public software package implemented in MATLAB was used to test PRGAMP. The other methods were implemented in Python with TensorFlow.} was compared only on the Gaussian model due to its structural characteristics. Even for the Gaussian model, PRED has a superior performance with lower complexity than existing algorithms including PRGAMP.\footnote{We excluded the performance result of PRGAMP in Fig. \ref{fig1j} because it is zero for the whole sparsity region.}

\section{Discussion} 

\subsection{The scalability of DNN to recover synthetic signals for SPRF}
\label{1discus}

To support the claim that the DNN structure can be scalable to estimate the support in SPRF, we prepared the following subsections 1) and 2). To show that the DNN is superior to other methods for solving SPRF, we prepared the following subsection 3). 

\smallskip \smallskip \smallskip  
\subsubsection{The DNN imposes structural priors for support estimation in SRPF}\label{discus1}

We introduce the following three results (a)--(c) by referring to \cite{schniter2015compressive}, \cite{he2017bayesian}, and \cite{liu2018low}:
\smallskip  
\begin{enumerate}[(a)] 
\item In \cite{he2017bayesian}, it is guaranteed that a canonical LSTM cell has the same structure as the computational flow of each iteration of Bayesian learning framework (BLF). 
\item In \cite{liu2018low}, it is shown that phase retrieval can be solved by using the BLF.
\item In \cite{schniter2015compressive}, the PRGAMP algorithm has an inner loop where the GAMP algorithm, one of compressed sensing algorithms for sparse linear inversion, is used to estimate the sparse signal. Thus, in PRGAMP, the GAMP algorithm can be replaced by any compressed sensing algorithm to estimate the support in SPRF. As the sparse BLF is a compressed sensing algorithm, we can conclude that SPRF can be solved by using the BLF; this also supports the result (b).
\end{enumerate}
\smallskip 
Results (a)--(c) imply that LSTM is a generalized (i.e., learned) version of the BLF (from result (a)) and support estimation in SPRF can be done by the BLF (from results (b) and (c)). Furthermore, it is well-known that the BLF is a scalable algorithm, as it has structural priors to estimate the target support. Therefore, as the BLF has structural priors, the LSTM-based DNN implicitly enables to impose structural priors to estimate the target support in SPRF. 
\smallskip \smallskip \smallskip  
\subsubsection{The DNN is scalable according to signal dimension n for support estimation in SRPF}\label{discus2}

Note that BLF is scalable for signal dimension $n$ and LSTM has the same structure as the BLF. Thus, LSTM can be scalable for $n$ by imposing the structure prior to estimate the target support in SPRF. Our test results shown in Figs. 1(a)--(c) implies that the LSTM-based DNN, used in the proposed PRED, estimates the true support with a high probability, irrespective of $n$. This is because PRED uniformly recovers about the twice the sparsity with a lower complexity than other related methods, for different values of $n$. This supports our claim that DNN (i.e., LSTM) is scalable for $n$ to estimate the target support in SPRF. 

Note that in our test, the signal is sampled from two continuous probability distributions (i.e., uniform and Gaussian). This implies that there is an infinite number of combinations of pairs for measurement vector $y$ and its corresponding true support, given any fixed sparsity $k$. Thus, it would not possible for the DNN to recover all true supports, given that the DNN should store all the infinite number of cases if the DNN did not have any inherent structure. 

The test results in Figs. 1(a)--(c) show that the DNN in PRED can recover twice the sparsity, by recovering all supports with probability one, in comparison with related SPRF methods. Hence, the DNN (i.e., LSTM) does not simply store all the supports. Instead, it has an implicit structure to estimate the target support via its forward computational flow, which is like a computational flow of the BLF. This inherent structure also ensures that the DNN is scalable.
\smallskip \smallskip \smallskip  
\subsubsection{The DNN can outperform existing methods for support estimation in SRPF}\label{discus3}

The LSTM can be interpreted as a learned version of the BLF. It has been shown in \cite{metzler2017learned} and \cite{ito2019trainable} that learned versions of approximated message passing and the iterative shrinkage thresholding algorithm outperform their counterparts for estimating the target support. This indicates that LSTM (i.e., the learned version of the BLF) can outperform the BLF for estimating the support in SPRF, as demonstrated in \cite{he2017bayesian} though by solving a problem different from SPRF. This implies from the results (b) and (c) in Section \ref{discus1} that DNN (i.e., LSTM) can outperform existing support estimators for SPRF. We demonstrated in Section \ref{sec_sim} that the proposed PRED outperforms other SPRF methods by using the LSTM as the DNN architecture for PRED, supporting our claim.  


\subsection{Demonstration of PRED scalability through intuition and principles}
\label{2discus}

PRED has the following two main steps (1) and (2): (1) Extended support estimation from DNN output and (2) support estimation from the extended support estimate via the TSE algorithm (Algorithm \ref{alg4}). For step (1), the DNN provides a set (the extended support estimate) including the true support with high probability, as the DNN (i.e., LSTM) has the implicit structure to estimate the support as discussed in Section \ref{1discus}. Step (1) has a low complexity as the extended support estimation using DNN is performed simply via a matrix multiplication at each DNN layer without solving specific optimization problems. Thus, the complexity (performance) of PRED is mainly dependent on step (2). As we have shown in the last paragraph of Section \ref{sec_sim}, the complexity of step (2) is O($b^2$), where b is the size of the extended support estimate obtained from step (1). Thus, the maximum of $b$ is $3k$ where $k$ is the sparsity. Hence, the complexity of PRED is the order of $k$ (i.e., O($k^2$)) and does not depend on signal dimension $n$. Disregarding step (1), the complexity of PRED is O($n^2$), as the extended support estimate in step (2) is set to the whole index set. Therefore, PRED is scalable (the complexity is not affected by $n$ but by $k$) by making the TSE algorithm search the support not in the whole index set, but in the extended support estimate obtained from the DNN. This justifies the combination of the DNN with existing SPRF algorithms, as discussed in Section \ref{conclusion}.

\section{Conclusion} \label{conclusion}
Although a DNN cannot accurately estimate the support, it is efficient to estimate the set containing it \cite{he2017bayesian}. On the other hand, the optimization-based approach is less efficient at finding the support from a full set of indices, but is highly accurate from a relatively small set including the support. We leverage the advantages of both approaches to perform DNN-based extended support estimation and first show that this approach, called PRED, outperforms existing algorithms in recovering common sparse signals for SPRF.

\appendix \label{dnntraining}
\section*{DNN Training} 
\subsection{Description of the proposed algorithm for training DNN}
Algorithm \ref{alg3} describes the proposed training method for the DNN $f_{\theta}(\cdot)$. It considers noisy training data for the DNN to estimate the UES, and the case when sparsity $k$ of ${x^{\circ}}$ in the test data is unknown, with minimum and maximum bounds given by $k_1$ and $k_2$, respectively, to generate the training data. In the algorithm, $n_{e}$ is the number of epochs, $s_b$ is the size of batch data, $n_{b}$ is the number of batches per epoch, and $v_\textup{SNR$_{\textup{dB}}$}$ is the SNR in decibels (dB), which is expected to be $10 \log_{10} \, (\sum_i c[i] / \sum_i w[i])$.
For each epoch, steps 3--9 generate training data $(x_i,y_i)_{i=1}^{s_b \cdot n_b}$, where $x_i$ and $y_i$ are the signal and measurement vectors, respectively. Specifically, in step 5, signal vector $x_i$, whose sparsity $s$ is uniformly sampled between $k_1$ and $k_2$, is sampled from the conditional probability $p_{{x^{\circ}}}(x \mid s)$ of the signal vector given sparsity $s$, defined as 
\begin{align}\nonumber
p_{{x^{\circ}}}(x \mid s):= \frac{p_{{x^{\circ}}}(x) \cdot 1(|\supp(x)|=s)}{\sum_{\tilde x \in \mathbb{R}^n \textup{ s.t. } |\supp(\tilde x)|=s}p_{{x^{\circ}}}(\tilde x)},
\end{align}
where $p_{{x^{\circ}}}(x)$ is the distribution of ${x^{\circ}}$ and $1(\cdot)$ is the indicator function. Measurement vector $y_i$ in step 9 is given by $z_i := F x_i$ plus noise vector $w_i$ such that $10 \log_{10} \, (\sum_j z_i[j] / \sum_j w_i[j]) \geq v_\textup{SNR$_{\textup{dB}}$}$. The training goal is to minimize cost $L(\theta):= \mathbb{E}_{(y,x)} [ \textup{ce(}f_{\theta}(y),\textup{u}_n(\mathcal{I})) ]$ in step 13, 
with $\textup{u}_n(\mathcal{I})$ being an $(n-1)$-dimensional vector $v \in \mathbb{R}^{n-1}$, whose nonzero elements are $1/|\mathcal{I}|$ and support is given by set $\mathcal{I}$, and $\textup{ce}(v_1,v_2):= -\frac{1}{g} \sum_{i=1}^{g} v_2[i] \log \, v_1[i] $ is the cross-entropy between vectors $v_1$ and $v_2$ of dimension $g$. Training parameter $\theta$ is updated to minimize $L(\theta)$ in steps 14 and 15 through $\textup{Update}_{\theta}(L(\theta), \eta)$ with its learning rate $\eta $. 
\begin{algorithm}[t]
\footnotesize
  \caption{TrainDNN(${F},k_1,k_2,s_b,n_b,n_e,v_\textup{SNR$_{\textup{dB}}$}$)}
    \begin{algorithmic}[1]
    \Input{${F} \in \mathbb{C}^{m \times n}$, $(k_1,k_2,s_b,n_b,n_e) \in \mathbb{N}^4$, $(v_\textup{SNR$_{\textup{dB}}$},\kappa) \in \mathbb{R}^2$}
    \Output{Neural network $f_{\theta}(\cdot): \mathbb{R}^{m} \rightarrow \mathbb{T}^{n-1}$}

 \For{$i \in \{1:n_{e}\}$}
    \For{$i \in \{1:n_{b} \cdot s_b\}$}
   \State $s  \in \mathbb{N} \gets$ integer uniformly sampled from  $\{k_1:k_2\}$
    \State $\mathcal{\bar T}_i \gets $  index set such with size $s$ and elements uniformly sampled from  $\{1:n\}$ 
    \State $x_i \in \mathbb{R}^{n}  \gets v \in \mathbb{R}^{n} \textup{ s.t. } \supp(v) = \mathcal{\bar T}_i $ and nonzero elements of $v$ are sampled from probability distribution $p_{{x^{\circ}}}(\cdot \mid s)$ of target signal vector ${x^{\circ}}$
\State  $\alpha \in \mathbb{R} \gets $  randomly sampled number from $0$ to $1$
\State $\bar w_i \in \mathbb{R}^{m} \gets $ randomly sampled from region satisfying $\sum_{i=1}^{n}\bar w[i]=1,\bar w[i]\geq 0$ for $i\in\{1:n\}$
\State $w_i \gets \frac{\alpha\cdot   \bar w_i }{ \sum_{j \in \{1:n\}} z_i [j]} \cdot 10^{- v_\textup{SNR$_{\textup{dB}}$}/10}$ where $z_i:=\abs(F x_i)^2$
\State $y_i \gets z_i+w_i$
\State $\mathcal{I}_i \gets \mathcal{I}_{-1}(\mathcal{\bar T}_i) -1$
    \EndFor

\For{$q \in   \{1:n_{b}\}$}
\State $L(\theta) \gets \frac{1 }{s_b}[\sum\limits_{p \in [(s_b-1)\cdot  q+1:s_b \cdot q]}\textup{ce(}f_{\theta}(y_p),\textup{u}_n(\mathcal{I}_p))]$
\State $\eta_q \gets \textup{set learning rate for gradient update}$
\State $\theta \gets \textup{Update}_{\theta}(L(\theta), \eta_q)$
\EndFor
\EndFor
\State \Return{$f_{\theta}(\cdot)$}
  \end{algorithmic}
\label{alg3}
\end{algorithm}

\subsection{Setting environment for the experiments in Section \ref{sec_sim}}
To train the network given by $f_\theta(\cdot)$, we used Algorithm \ref{alg3} by setting input $(k_1,k_2,s_b,n_b,n_e)$ to $(2,20,10^6,250,40)$ and fixing $v_{\textup{SNR$_{\textup{dB}}$}}$ to the SNR. In addition, we used RMSprop optimization with learning rate $\eta_i$ of $0.0001$ for epochs $i \leq 10$ and $0.0001/4^j$ for epochs $i$ from $1+10j$ to $10+10j$ ($j \in \{1:3\}$) to update the gradient.

\balance
\bibliographystyle{IEEEtran}
\bibliography{IEEEabrv,bibdata4}

\ifCLASSOPTIONcaptionsoff
  \newpage
\fi

\end{document}